\documentclass[electronics,article,accept,moreauthors,pdftex]{Definitions/mdpi} 
\firstpage{1} 
\pubvolume{9}
\issuenum{6}
\articlenumber{936}
\pubyear{2020}
\copyrightyear{2020}
\history{Received: 9 May 2020; Accepted: 1 June 2020; Published: 4 June 2020}
\updates{yes} 

\pdfoutput=1

\usepackage{algorithm}
\usepackage{algorithmic}
\usepackage{graphicx}       % Required
\usepackage{amsmath}
\usepackage{multirow}
\usepackage{color}
\usepackage{wrapfig}

\Title{DeepSoCS: A Neural Scheduler for Heterogeneous System-on-Chip (SoC) Resource Scheduling}

\Author{Tegg Taekyong Sung $^{1,}$*$^{,\dagger}$\orcidA{}, Jeongsoo Ha $^{2,\dagger}$\orcidB{}, Jeewoo Kim $^{3,\dagger}$\orcidC{}, Alex Yahja $^{4}$\orcidD{}, \mbox{Chae-Bong Sohn $^{1,}$*\orcidE{}} and Bo Ryu $^{4}$\orcidF{}}
\AuthorNames{Tegg Taekyong Sung, Jeongsoo Ha, Jeewoo Kim, Alex Yahja, Chae-Bong Sohn and Bo Ryu}

\address{%
$^{1}$ \quad Department of Electronics and Communications Engineering, Kwangwoon University, Seoul 01897, Korea\\
$^{2}$ \quad Department of Computer Science and Engineering, Chungnam National University, Daejeon 34134, Korea; jeongsoo.ha1210@gmail.com\\
$^{3}$ \quad Department of Computing, Imperial College London, London SW7 2AZ, UK; jeewoo.kim17@imperial.ac.uk\\
$^{4}$ \quad EpiSys Science, Poway, CA 92064, USA; alex@episyscience.com~(A.Y.); boryu@episyscience.com~(B.R.)}

\corres{Correspondence: tegg89@gmail.com~(T.T.S.); 
cbsohn@kw.ac.kr~(C.-B.S.)}

\firstnote{Work done while at Episys Science, Poway, CA 92064, USA.} 
\abstract{In this paper, we~present a novel scheduling solution for a class of System-on-Chip (SoC) systems where heterogeneous chip resources (DSP, FPGA, GPU, etc.) must be efficiently scheduled for continuously arriving hierarchical jobs with their tasks represented by a directed acyclic graph. Traditionally, heuristic algorithms have been widely used for many resource scheduling domains, and Heterogeneous Earliest Finish Time (HEFT) has been a dominating state-of-the-art technique across a broad range of heterogeneous resource scheduling domains over many years. Despite their long-standing popularity, HEFT-like algorithms are known to be vulnerable to a small amount of noise added to the environment. Our Deep Reinforcement Learning (DRL)-based SoC Scheduler (DeepSoCS), capable of learning the ``best'' task ordering under dynamic environment changes, overcomes the brittleness of rule-based schedulers such as HEFT with significantly higher performance across different types of jobs. We~describe a DeepSoCS design process using a real-time heterogeneous SoC scheduling emulator, discuss major challenges, and present two novel neural network design features that lead to outperforming HEFT: (i) hierarchical job- and task-graph embedding; and (ii) efficient use of real-time task information in the state space. Furthermore, we~introduce effective techniques to address two fundamental challenges present in our environment: delayed consequences and joint actions. Through an extensive simulation study, we~show that our DeepSoCS exhibits the significantly higher performance of job execution time than that of HEFT with a higher level of robustness under realistic noise conditions. We~conclude with a discussion of the potential improvements for our DeepSoCS neural scheduler.}

\keyword{heterogeneous resource scheduling; deep reinforcement learning; neural networks; system on chip}

\begin{document}
%%%%%%%%%%%%%%%%%%%%%%%%%%%%%%%%%%%%%%%%%%

%\maketitle

%\input{intro}
\section{Introduction}
\label{sec:introduction}
\textls[-31]{Task scheduling is a universal problem that affects many aspects of our lives, including wireless communication systems, supply chain logistics, device placement, computer processors, supercomputing, and~cloud computing, to~name a few. Any algorithm achieving higher resource-efficient task/job execution without creating an additional system penalty can bring huge benefits, lower costs, or~both, to~many industries. To~date, heuristic-based list scheduling algorithms are widely used in a multitude of heterogeneous task and resource scheduling problems, where they heuristically search relative importance in presented task nodes and schedule the next task on the rank basis. Heterogeneous Earliest Finish Time (HEFT) is a general list scheduler showing the state-of-the-art performance~\cite{topcuoglu2002performance,beaumont2019survey}. HEFT and its derivative Predict Earliest Finish Time (PEFT)~\cite{arabnejad2013list} are thus primary benchmarks to compare against. To~this date, these algorithms both generate competitive scheduling decisions in the context of minimizing total application execution time~\cite{maurya2018benchmarking}.}

Most heuristic algorithms need handcrafted rules, and~therefore, are difficult to adapt to other domains without significant and time-consuming design changes, especially in complex and dynamic systems. But~perhaps their most significant drawback is that it is susceptible to even a small amount of noise presented in the environment, often leading to significantly degraded performance. To~overcome these limits, we~have investigated a Deep Reinforcement Learning (DRL) based approach that is capable of learning to schedule a multitude of jobs without significant design changes while simultaneously addressing the inherently high brittleness of rule-based schedulers with higher system-wide performance. In~particular, our algorithm learns to schedule hierarchical job-task workloads for heterogeneous resources such as system-on-chip (SoC) processors with extremely stringent real-time performance~constraints.

\textls[-15]{DRL enables a trainable agent to learn the best actions from interactions with the environment. DRL based algorithms have achieved human-level performance in a variety of environments, including video games~\cite{mnih2015human}, zero-sum games~\cite{silver2016mastering}, robotic grasping~\cite{kalashnikov2018scalable}, and~in-hand manipulation tasks~\cite{andrychowicz2020learning}. There~have been many solutions proposed for a variety of task scheduling applications. One such scheme is Decima, a~combined graph neural networks and actor-critic algorithm, which has demonstrated its capability to successfully learn to schedule hierarchical jobs for cloud computing resources with high efficiency~\cite{mao2018learning}. However, Decima is not directly applicable to our SoC processor scheduling domain for the following two reasons. First, the~job injection rate of Decima is kept very low with virtually no job overlapping, whereas, in~a real-world SoC system, the~job injection rate may be much higher with a reasonable degree of overlapping. Second, while the objective of Decima is to achieve the shortest execution time of scheduling a predefined number of jobs, the~goal of our scheduler is to complete as many jobs as possible in a given time with no predefined number of jobs as a target. Understanding these stark differences present in our SoC environment is essential to develop a new, practical, and~high-performance scheduler for heterogeneous SoC applications that differentiates itself from the class of Decima~schedulers.}

In addition to recognizing the differences between the Decima and our scheduler design environments (cloud computing vs. SoC processors), it is also critical to address new challenges that stem from utilizing high-fidelity simulators used by SoC designers to represent the environment. To~develop a practical SoC resource scheduler, it is imperative to use highly realistic simulators (e.g., Discrete-event Domain-Specific System-on-Chip Simulation, or~DS3) used by a broad SoC design community~\cite{arda2020ds3}. As~reported in prior work~\cite{dulac2019challenges}, the~use of real-world environments for DRL design such as DS3 often comes with steep costs. For~example, the~reward corresponding to the agent's actions is often not immediately received by the DRL agent when running inside real-world simulators. Known as a delayed consequence, this poses a substantial challenge in reward shaping due to the unpredictable nature of the delays. Also, it is challenging for the agent to fully grasp the environment state in real-time, which leads to partial observability problem and the associated state representation design challenge. Furthermore, the~scheduler must perform actions for every task in the task queue that its choices are dynamically changed every time step, resulting in policy optimization~challenge.

To address these challenges, we~introduce DeepSoCS, a~novel neural network algorithm that learns to make the extremely resource-efficient task ordering actions in a reward-delayed, concurrent, real-time task-execution environment. We~evaluate the performance of the DeepSoCS through an extensive simulation study and using real-world SoC simulator to demonstrate the robustness and system-wide performance gains in job execution time under both realistic noise and noise-free conditions over HEFT. To~the best of our knowledge, DeepSoCS is the first neural scheduler that outperforms HEFT in a general heterogeneous system-on-chip (SoC) scheduling~domain. 
 
The rest of the paper is organized as follows.  Section~\ref{sec:prob-scenario} introduces the real-world DS3 simulation tool (widely used by SoC chip design researchers and engineers) and its challenging constraints that impact our design. Section~\ref{sec:proposed-method} describes the overall DeepSoCS architecture and its two novel techniques aimed at addressing the delayed consequence and joint-action problems. Section~\ref{sec:experiments} shows experimental results that compare DeepSoCS to HEFT. Section~\ref{sec:related-work} describes related works of the job scheduling problem. Finally, Section~\ref{sec:conclusion} provides the conclusions and future research~directions.

%%%%%%%%%%%%%%%%%%%%%%%%%%%%%%%%%%%%%%
%\input{scenario}
\section{Problem~Scenario}
\label{sec:prob-scenario}
The objective of scheduling algorithms is makespan minimization. The~optimal scheduler must find the best mapping from the tasks to the processors (processing elements or PEs) given a task graph represented by a Directed Acyclic Graph (DAG) and a set of heterogeneous computing resources. In~most practical situations, makespan minimization is NP-hard~\cite{shirazi1995scheduling}. The~heuristic algorithms typically need handcrafted rules and, especially, are vulnerable to noise and changes in an environment, which can lead to a significant reduction in performance. To~build a scheduler with robustness to dynamic changes and noises in the real world, we~adopt a learning-based algorithm.
In this section, we~introduce the structure of DS3 simulator designed for heterogeneous resource scheduling to give a better understanding of agent and environment interactions. Furthermore, the~fundamental challenges of DRL in a realistic simulation is discussed~\cite{dulac2019challenges}.

\subsection{DS3~Simulation}
\label{sec:prob-def:simulation}
\begin{figure}[H]
  \centering
    \includegraphics[height=0.35\linewidth]{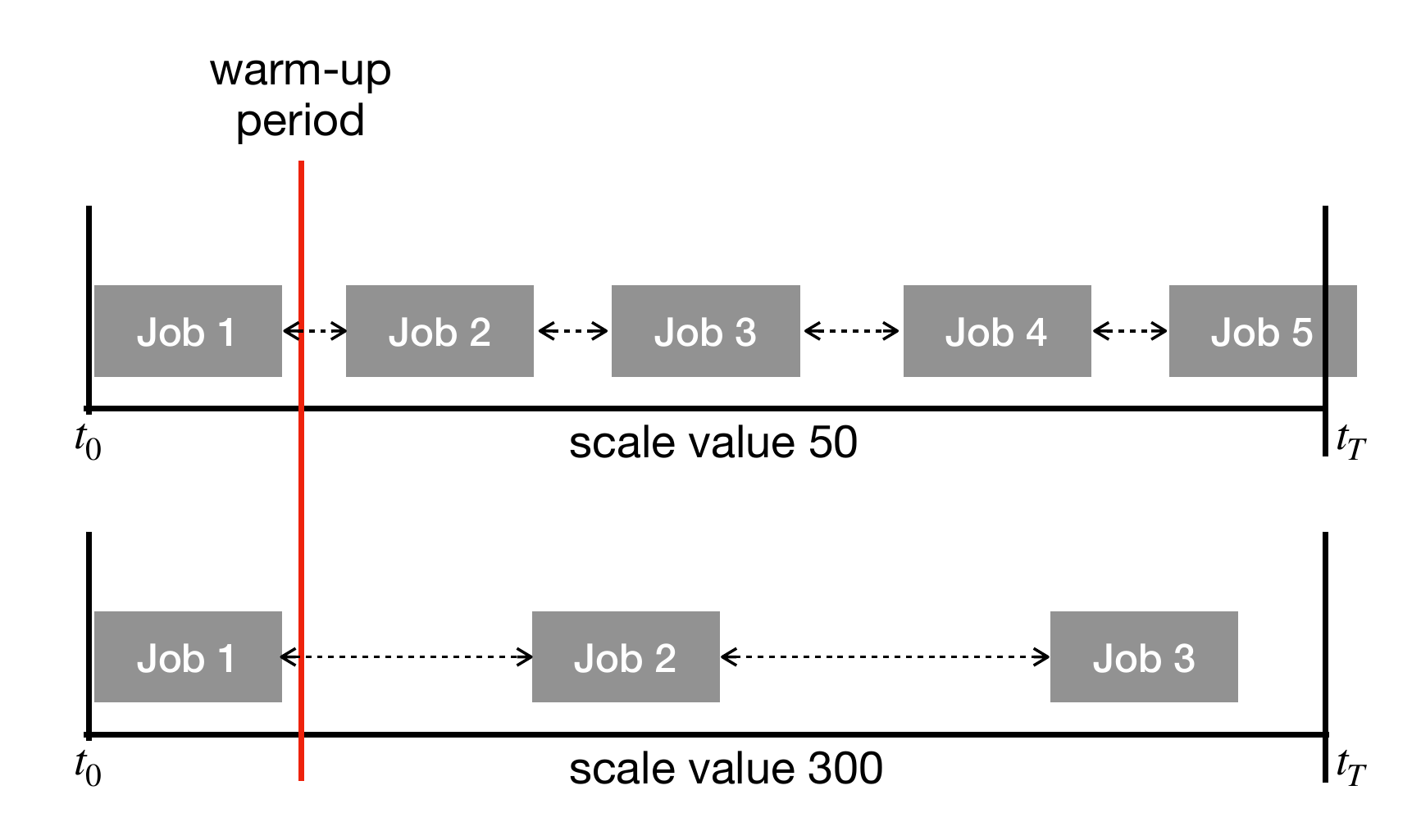}
    \caption{A timeline of multiple jobs injected into the job queue at different scales. A~smaller scale value means a higher injecting rate. Fast job injections make scheduling tasks more difficult, especially~when there are continuously overlapping jobs, which add more complexity. The~timeline starts at $t_0$ and finishes at $t_T$, where $T$ is a final time step. Note that in the top diagram, job 5 is discarded as the simulation terminates, because~the performance only takes account of completed~jobs.}
    \label{fig:jobgen-timeline}
\end{figure}

A discrete-event Domain-Specific System-on-Chip Simulation (DS3) is a real-time system-level emulator that is built for scheduling tasks to general-purpose and special-purpose processors, especially optimizing the processors to a particular domain~\cite{arda2020ds3}. It is known as domain-specific system-on-chips (DSSoCs), a~class of heterogeneous architectures. It allows users to develop algorithms on run-time and explore algorithms rapidly, and~also provides built-in table-based schedulers and heuristic algorithms as baselines. The~overall system of DS3 is shown in Figure~\ref{fig:jobgen-timeline}. The~jobs are continuously injected into the job queue at every $t$ time step, where $t \sim \text{Exp}(\frac{1}{\text{scale}})$. The~scale value, which controls a job injection rate, is given by the simulator. Throughout the paper, we~consider non-preemptive and steady-state scheduling~\cite{beaumont2005steady}. The~environment provides a `warm-up period' so that the simulation can start at the steady-state. The~simulation discards any results not reached to the steady-state. Our objective is to complete as many jobs as possible within a given simulation length. Faster job execution means more jobs can be injected into the job queue in the simulation, due to the capacity of the job queue. Therefore, the~evaluation criteria is an average latency, where $latency = \frac{total\; exec\; time}{total\; completed\; jobs}$. 

\begin{figure}[H]
  \centering
    \includegraphics[height=0.4\linewidth]{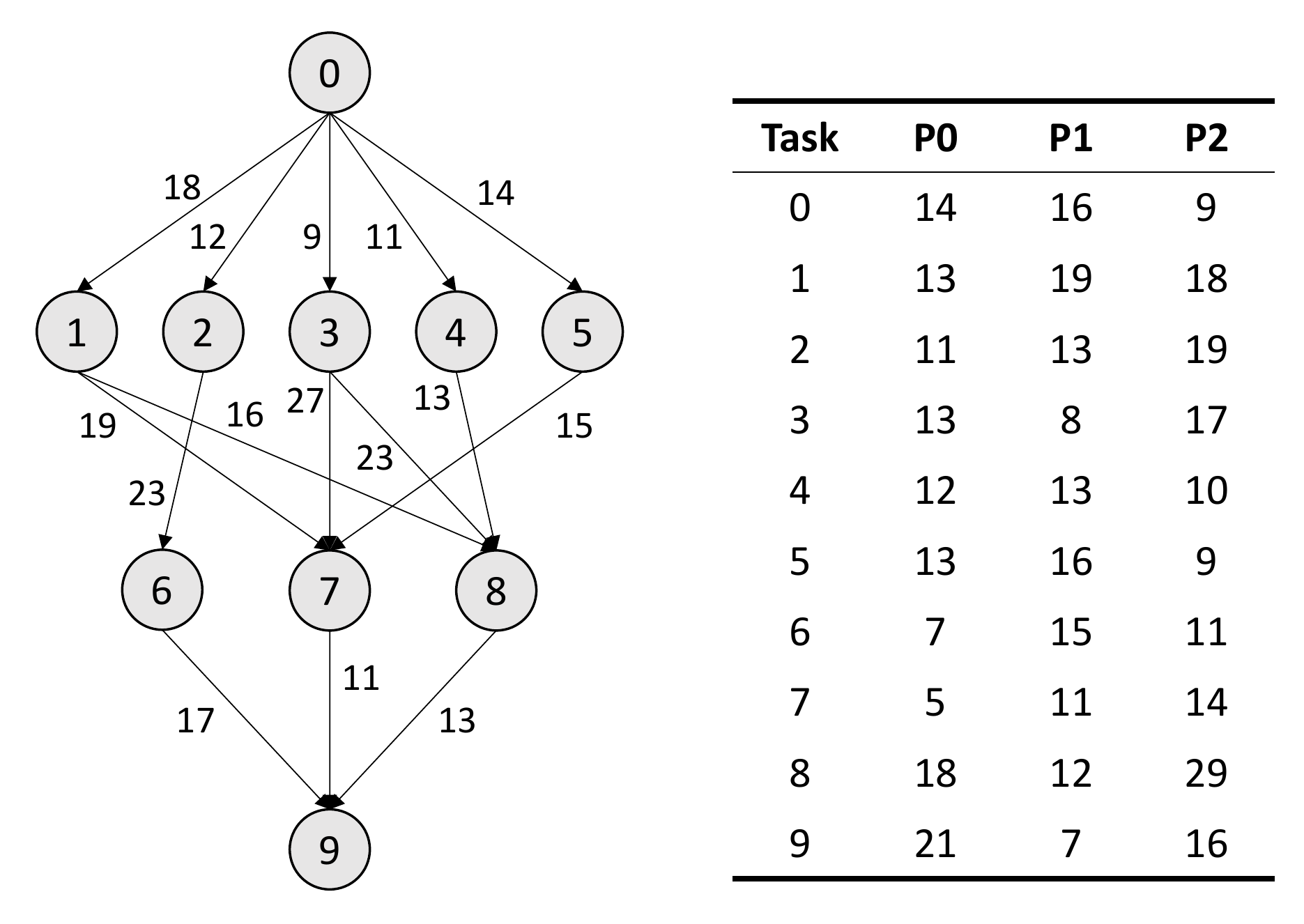}
    \caption{A diagram showing a canonical job and resource profiles~\cite{topcuoglu2002performance}. This~job consists of 10 tasks, and~3 resources support all tasks. On~the left figure, nodes represent tasks and edges represent communication costs. The~right table describes execution time for supporting functionalities on each processor. A~more complicated WiFi profile is described in Appendix~\ref{sec:appendix:wifi}.}
    \label{fig:top-profile}
\end{figure}

\begin{figure}[H]
  \centering
    \includegraphics[height=0.4\linewidth]{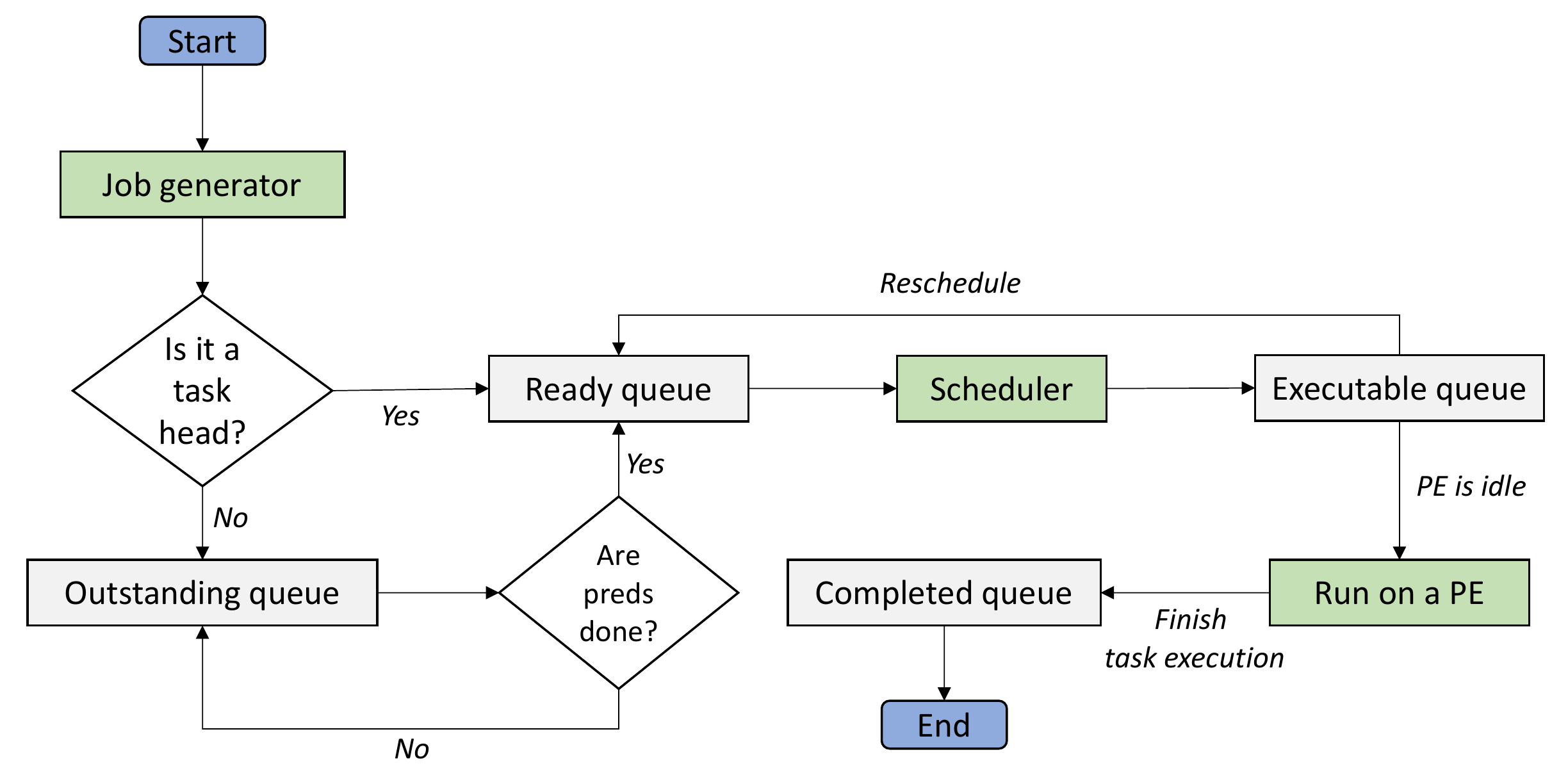}
    \caption{The DS3 life-cycle from job generation to task execution. First, the~job generator injects a job to the job queue, and~its tasks are loaded to the corresponding task queues. Then, the~scheduler selects tasks in the ready queue, and~assigns them to PEs, and~the idle PEs run the scheduled tasks. Any task remained in the executable queue can be reloaded to the ready queue and rescheduled. Once the scheduled task is completed, it is moved to the completed~queue.}
    \label{fig:ds3-lifecycle}
\end{figure}

The input job is represented as a DAG structure where each node represents a specific task. Figure~\ref{fig:top-profile} shows an example of a canonical job DAG and resource profiles~\cite{topcuoglu2002performance}. In~a single job, each task is structured with a task dependency graph, and~a scheduler only assigns the tasks with no predecessors or the tasks which its predecessors are all completed. The~edges represent communication costs computed from one processing element to another processing element (PE). Each processor supports functionalities, and~their task execution time is listed on the right in Figure~\ref{fig:top-profile}. In~this profile, the~best mappings for the first two tasks are T0 to P2 and T1 to P0, where T is a task, and~P is a processor. The~tasks scheduled to the processors currently executing the task remains in the executable queue until the processor becomes idle. Simulation with multiple jobs adds complexity. When the designated input profiles are loaded in the DS3 system, jobs are continuously injected into the job queue by job generator, and~the corresponding tasks are loaded to the task queue. Then the tasks follow the DS3 life cycle as described in Figure~\ref{fig:ds3-lifecycle}. 

\unskip

\unskip

\subsection{Challenges}
\label{sec:prob-def:challenges}
\begin{figure}[H]
    \centering
    \includegraphics[width=0.6\textwidth]{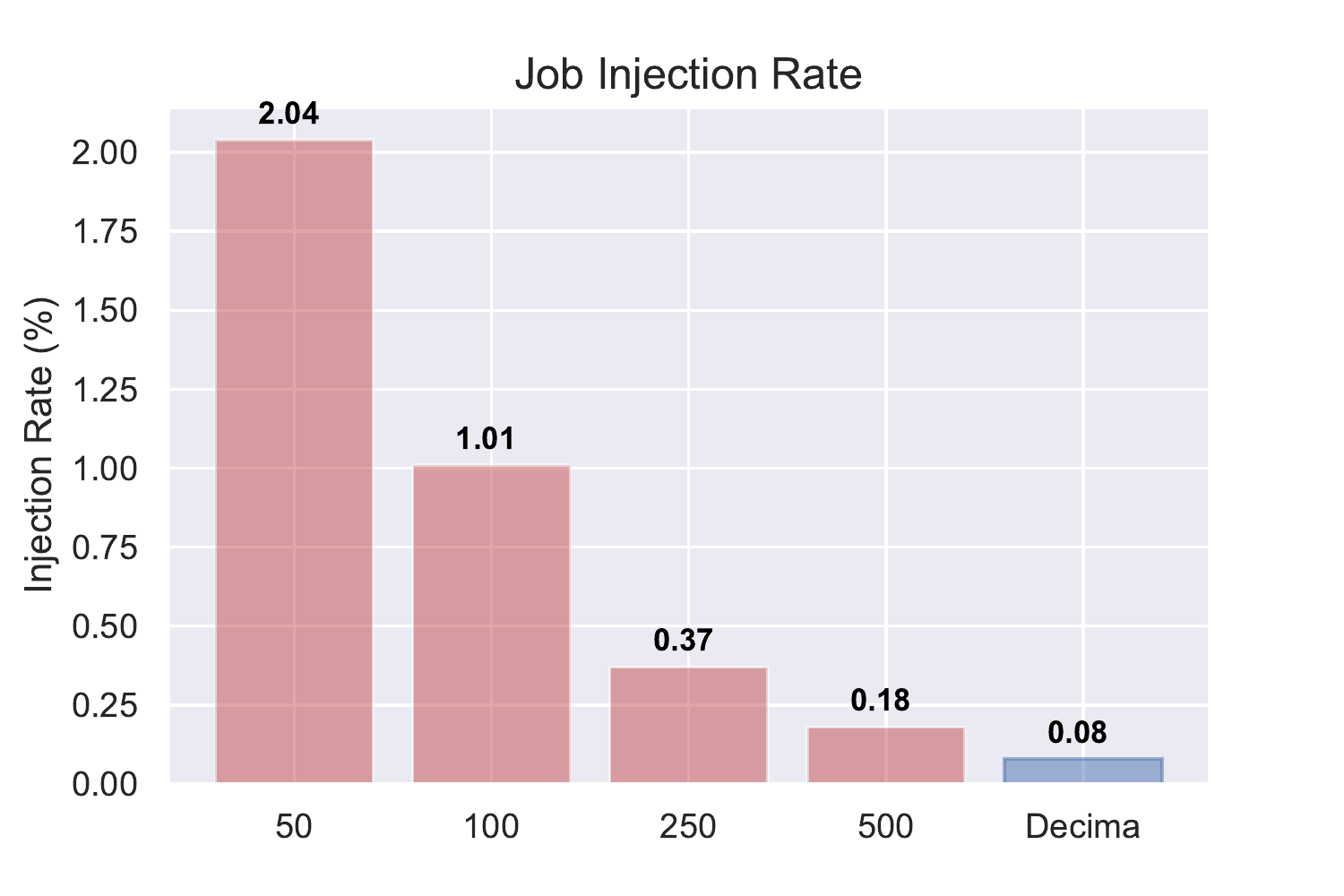}
    \caption{\textls[-15]{A job injection rate comparison of DS3 and Decima. The~rightmost box shows a job injection rate of Decima, and~the other boxes show job injection rates of DS3 with different scale values. As~described in the plot, DS3 can simulate with a significantly high injection rate, and~especially input jobs significantly overlapped on the scale of 50. For~Decima, we~used default parameters following the~paper.}}
    \label{fig:job-load}
\end{figure}

Some of the recent studies attempt to use learning-based algorithms in task scheduling domains. For~example, Decima uses hierarchical and heterogeneous jobs to homogeneous executors and schedules tasks based on a continuous-time frame~\cite{mao2018learning}. However, Decima pre-defines the number of jobs, and the injection time step, and~the job injection rate is significantly lower than DS3, as~shown in Figure~\ref{fig:job-load}. In~many real-world systems, jobs overlapping due to high injection rate and endless job generation is often the reality. Contrary to the environment used in Decima, DS3 continuously generates jobs until the termination of the simulation without any predefined information. Therefore, the~objective is to complete as maximal completion of jobs with the shortest~time.

Next, we~investigate two main challenges when applying the RL agent to the DS3 environment. First, Markov Decision Process (MDP) is violated due to the asynchronous transitions between the agent and the environment. The~DS3 environment operates in real-time. A~state is observed whenever the tasks are inserted into the ready queue. Also, the~agent must take actions for every task in the ready queue. The~rewards from these actions are not calculated until the completion of the assigned tasks, causing the delayed reward. But, before~the reward is calculated, the~subsequent tasks of the previously executed tasks arrive at the ready queue, and~the agent retakes actions. As~this is repeated throughout the simulation, the~transition elements are collected asynchronously, which results in an MDP violation. Second, due~to the mechanism of DS3, that it orders all the tasks in the ready queue and assigns them to PEs, and~task dependencies, the~agent's action space changes at every timestep, resulting in a combinatorial optimization problem. Furthermore, it brings credit assignment problems, where the agent tries to maximize the long-term goal of the maximum number of completed jobs with the shortest time. The~above difficulties remain as open~problems. 

%%%%%%%%%%%%%%%%%%%%%%%%%%%%%%%%%%%%%%
%\input{proposed}

\section{Proposed~Method}
\label{sec:proposed-method}
In this section, we~introduce our newly proposed architecture called DeepSoCS, which applies deep reinforcement learning (DRL) to learn the best task ordering under dynamic environment changes. DeepSoCS is composed of PE manager, which maps tasks to PEs, and~task manager, which adaptively orders input tasks. We~design our DRL algorithm to overcome the limitations of existing DRL algorithms in the real world: partial observability, stochastic dynamics of the environment, sparse reward functions, and~unknown delays in the system's actions or rewards~\cite{dulac2019challenges}. Furthermore, we~discuss two main challenges that arise from the realistic environment DS3: (i) delayed responses to an action (ii) joint~action.
 
\subsection{PE~Manager}
\label{sec:proposed-method:pe}
Both DeepSoCS and HEFT follow the Earliest execution Finish Time (EFT) algorithm, which~heuristically maps the available PEs to the ordered tasks based on communication and computation costs. The~EFT algorithm is introduced in the ``List Scheduling'' domain and is based on the Earliest execution Start Time (EST) algorithm~\cite{topcuoglu2002performance}. The~EST is initialized to 0 for the entry task node, $\text{EST}(n_{entry}, p_j) = 0$. Then the EST recursively computes values starting from the entry task, as~shown in Equation~\eqref{eq:est}.
\begin{equation}
\label{eq:est}
    \text{EST}(n_i, p_j) = \max \Big\{ \text{avail}[j], \max_{n_m \in pred(n_i)} \big( \text{AFT}(n_m) + c_{m,i} \big) \Big\},
\end{equation}
where $n_i$ is task $i$, $p_j$ is processor $j$, $\text{avail}[j]$ is the earliest time at which processor $p_j$ is ready for executing the task, $pred(n_i)$ is the set of immediate predecessor tasks of task $n_i$, \text{AFT} is the actual finish time, and~$c_{m,i}$ is communication time from $t_m$ to $t_i$. 

Then, the~EFT algorithm is formalized by adding average execution cost, $w_{i,j}$, as~shown in Equation~\eqref{eq:eft}.
\begin{equation}
\label{eq:eft}
    \text{EFT}(n_i, p_j) = w_{i,j} + \text{EST}(n_i, p_j),
\end{equation}
where $w_{i,j}$ is the execution time to complete task $t_i$ on processor $p_j$. The~EFT algorithm here also has an insertion-based policy that considers the possible insertion of a task in an earliest idle time slot between two already-scheduled tasks in their slots on a~processor.

\subsection{Task~Manager}
\label{sec:proposed-method:task}
\textls[-15]{It is essential to efficiently order tasks first because PE is greedily selected with respect to the task ordering. The~baseline algorithm, HEFT, uses $rank_u$ value as a criterion of the task order. The~$rank_u$ value is computed with the task computation costs and the communication costs from available tasks. It represents the length of the critical path from task $i$ to the exit task. $rank_u(n_j) = \overline{w_i} + \max_{n_j \in succ(n_i)}(\overline{c_{i,j}} + rank_u(n_j))$, where $n_i$ represents the task $i$, $succ(n_i)$ is the set of immediate successors of task $i$, $\overline{c_{i,j}}$ is an average communication cost of task $i$ to task $j$, and~$\overline{w_i}$ is an average computation cost of task $i$. $rank_u$ values of all tasks are initially set to 0 and are recursively computed starting from the exit task by traversing the task graph reversely. Contrary to the HEFT which makes task orders by pre-computed $rank_u$ values, DeepSoCS uses a novel deep reinforcement learning method to adaptively prioritize input~tasks.}

% \mathcal{S} and S are different symbols. So we would like to keep those as-is.
In reinforcement learning, a~learning system can be modeled as a Markov Decision Process (MDP) with discrete time steps. Mathematically, the~MDP setting can be formalized as a 5-tuple $\langle \mathcal{S}, \mathcal{A}, R, \mathcal{P}, \gamma\rangle$~\cite{sutton2018reinforcement,puterman2014markov}. Here, $\mathcal{S}$ denotes the state space, $\mathcal{A}$, the~action space, and~$R: \mathcal{S} \times \mathcal{A} \times \mathcal{S} \rightarrow \mathbb{R} $, a~reward function which is defined over the state-action pair and the next state. $\mathcal{P}$, a~stochastic matrix specifying transition probabilities to next states given the state and the action, and~$\gamma \in [0, 1]$, a~discount factor. The~agent interacts with the environment and returns a trajectory $(S_1, A_1, R_1, S_2, ...)$, where~$S_{t+1} \sim \mathcal{P}(\cdot \mid S_{t}, A_{t})$. We~denote random variables in upper-case, and~their realizations in lower-case. MDP has the Markov property, defined as the independence of the conditional probability distribution of the future states of the process from any previous state, with~the exception of the current state. This~implies that the transitions only depend on the current state-action pair and not on the past state-action pairs nor on the information excluded from $s \in \mathcal{S}$. The~goal of the learner is to find an optimal control policy $\pi^\ast : \mathcal{S} \rightarrow \mathcal{A}$ that maps states to actions and that maximizes, from~every initial state $s_0$, the~return, i.e.,~the cumulative sum of discounted rewards $R$: $R(S_0) = \sum^\infty_{t=0} \gamma^t R_{t+1}$.

\begin{figure}[H]
  \centering
    \includegraphics[height=0.55\linewidth]{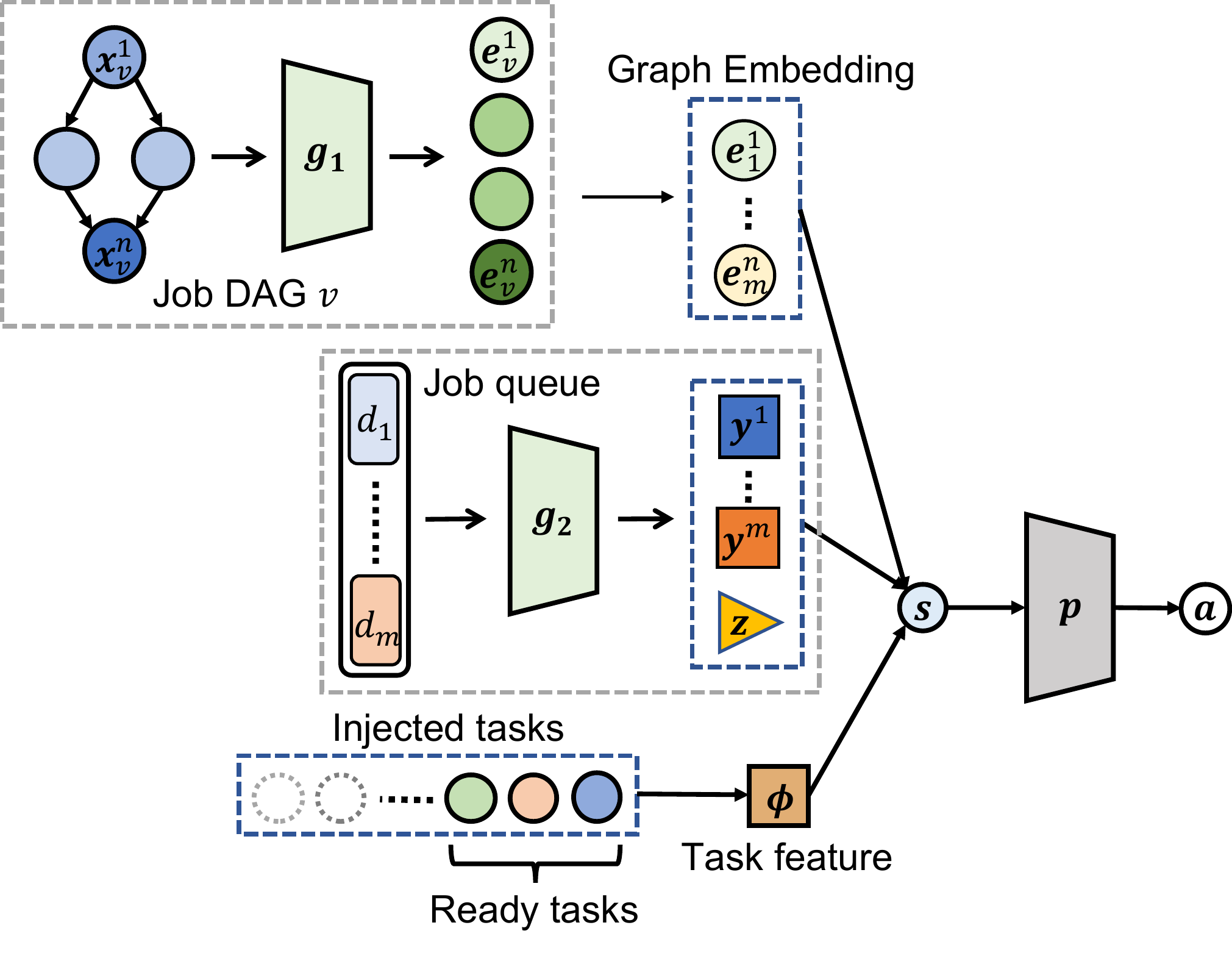}
    \caption{The task ordering is trained via DeepSoCS architecture. The~state is composed of graph embeddings and task features. A~node-level MPNNs, $g_1$, computes embedding nodes for each job injected in the job queue, and~a job-level MPNNs, $g_2$, computes local and global summaries using node embeddings and injected jobs information. Then, the~onward task information constructs task features, which represent the number of possible actions. We~use conventional policy networks $p$ to select a task. All vectors have time step subscripts but were not displayed in this diagram for~readability.}
    \label{fig:DeepSoCS-arch}
\end{figure}

Figure~\ref{fig:DeepSoCS-arch} describes the overall DeepSoCS networks structure. Two consecutive MPNNs~\cite{gilmer2017neural}, a~type of graph neural networks, inherently capture the important features of DAG structured jobs, such as task dependencies and communication costs. A~node-level MPNN, denoted as $g_1$, takes a job DAG as an input and computes task node features by grasping the features of its neighbor edges, which~we call as graph embeddings. A~job-level MPNN, denoted as $g_2$, takes all node features and injected jobs as inputs, and~computes the local feature of each job graph and the global feature of overall jobs. A~MPNN is $\textbf{e}_v = g \large[ \sum_{w \in \xi(v)} f(\textbf{e}_w) \large] + \textbf{x}_v$, where $f(\cdot)$ and $g(\cdot)$ are non-linear transformations, and~$\xi(v)$ refers to the set of $v$'s children. In~an individual injected DAG, $G_v$, its node, $\textbf{x}^i_v$, have aggregated messages from all their children nodes and computes its embedding, $\textbf{e}^i_v$ by a node-level MPNNs, $g_1$. Then, each node with its node embedding outputs DAG summary, $\textbf{y}^i$ and a global summary across all DAGs, $\textbf{z}$, with~a job-level MPNNs, $g_2$. Next, we~create normalized task features, $\phi$, denoting such information: PE statuses, DAG running identifier, running task duration, and~number of remaining tasks. The~task feature carries sufficient information since they are dynamically updated whenever the task is scheduled or PE executes the task. The~graph embedding and the task feature are concatenated to construct state, $\textbf{s} = \large[ \phi \| \{  \textbf{e}^1_1, \dots, \textbf{e}^n_m \} \| \{ \textbf{y}^1, \dots, \textbf{y}^m \} \| \textbf{z} \large]$. We~omit the time step for~legibility.

We use conventional policy networks to select actions, $\textbf{a}$, with~respect to its policy, $\pi_\theta (\textbf{s}, \textbf{a})$ defined as the probability of taking action $\textbf{a}$ in state $\textbf{s}$. The~cost can be computed using well-known actor-critic algorithm~\cite{konda2000actor}:
\begin{equation}
\label{eq:ac-cost}
    \nabla_\theta J(\theta) = \mathbb{E}_\pi \Big[ \sum_{t=1}^T \nabla_\theta \log \pi_\theta(a_t|s_t) \big( \sum^T_{t'=t} r_{t'} - b_t \big) + \beta H(\pi_\theta(\cdot|s_t)) \Big],
\end{equation}
where $H$ is entropy for the policy $\pi$, computed by $H(\pi_\theta(\cdot|s_t)) = \mathbb{E}_{a \sim \pi_\theta(\cdot|s_t)} [-\log \pi_\theta (a|s_t)]$, $\beta$ is a scaling factor, and~$b_t$ is a baseline used to reduce the variance of the estimated gradient. The~objective is to maximize the above cost function, and~the entropy regularizes the cost, resulting in exploration. $\beta$~is a hyperparameter, and~it is initially set to 1 and decays by 1e-3 every episode. An~actor network selects an action with respect to the policy, and~a critic network computes the baseline to reduce variances, $b_t = \mathbb{E}_{a_t \sim \pi_\theta}[Q(s_t, a_t)]$, where $b_t$ is a baseline at $s_t$. The~policy makes decisions based on the scheduling system and job arrival process, and~therefore we use ``input-dependent'' baselines to customize for different job arrival sequences~\cite{mao2018variance}. $\sum^T_{t'=t} r_{t'} - b_t$ estimates how much better the total reward is compared to the average reward in a particular episode. $\nabla_\theta \log \pi_\theta(a_t|s_t)$ provides a direction to increase the trajectory probability at action $a_t$ and state $s_t$.

In DS3 simulation, the~agent needs to schedule tasks and, consequently, completes as many jobs as possible within a reasonably long simulation length. We~consider the problem as an undiscounted \textls[-15]{infinite-horizon setting and therefore apply differential reward~\cite{sutton2018reinforcement} (\S10.3, \S13.6). Reward is a calculation} of the duration of all processing jobs.
\begin{equation}
\label{eq:mean}
\begin{split}
&  R_t = -C \times \sum^J_j (ct_j - st_j), \\
&  C = 
   \begin{cases}
     0,& \text{if } \sharp\text{completed jobs} = 0 \\
     \frac{1}{\sharp\text{completed jobs}},& \text{otherwise}
   \end{cases}
\end{split}
\end{equation}
where $J$ is a total number of injected jobs when invoking schedule function, $ct_j$ is the last completed time of job $j$, $st_j$ is injected time of job $j$. The~remaining job duration is continuously updated in every environment time step. When the ready task is not replenished in the ready queue, we~consider the agent taking a ``no-op'' action and recalculate the reward and update it to the reward storage. Although~the agent action is not completed (PE execution is on-going), the~agent receives the rewards at every time step because the reward calculates the ongoing job processes. Moreover, DS3 evaluation can be varied by setting different scale values. An~environment with low scale value (higher injection rate) is more complex to solve and lead to a bad evaluation. That being said, it is ideal for taking a cascade problem, so we train the agent via curriculum learning by gradually decreasing scale values~\cite{bengio2009curriculum}. 

\subsection{Delayed~Consequences}
\label{sec:proposed-method:delayed-outcome}
The delayed consequence is one of the fundamental challenges in RL~\cite{rlblogpost,dulac2019challenges}, and~often appears in real-time environments. MDP~\cite{bellman1957markovian} theoretically underpins conventional reinforcement learning (RL) methods and is well suited to represent turn-based decision problems such as board games (e.g., Go and Shogi). On~the other hand, it is ill-suited for real-time applications in which the environment state continues evolving dynamically without waiting for the agent's consideration and completing execution of an action~\cite{travnik2018reactive} such as task scheduling in our DS3 real-time system emulator. MDP could still be used in real-time applications by using some tricks, e.g.,~ensuring that the time required for action selection is nearly zero~\cite{hwangbo2017control} or pausing a simulated environment during action selection. Both of these, however, are not safe assumptions to make for mission-critical real-world~applications.

\begin{figure}[H]
  \centering
    \includegraphics[height=0.42\linewidth]{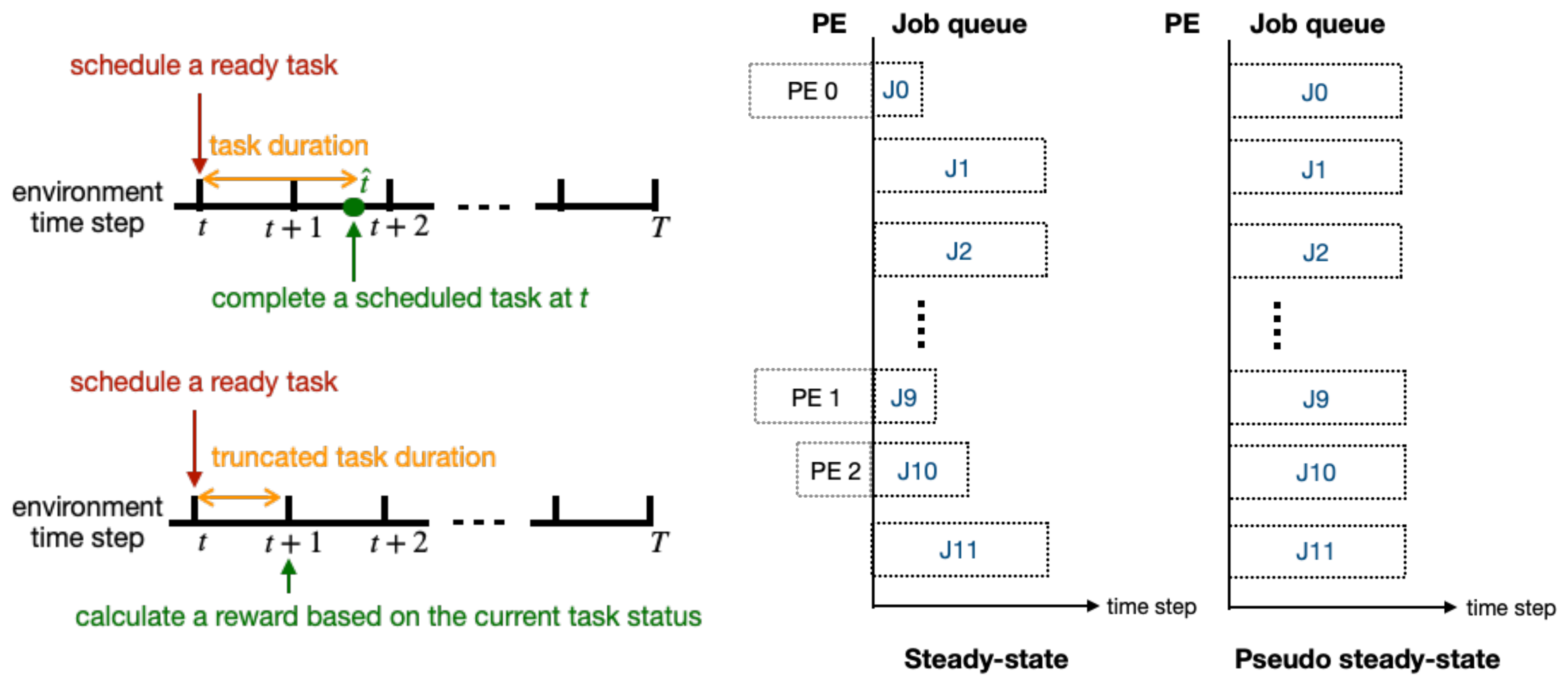}
    \caption{Left panel a timeline for the agent-environment interaction. The~top figure illustrates that the reward is received after the scheduled task is completed. We~emphasize that the previously scheduled task has not completed yet, but~the agent receives the next state because any task with no predecessor can arrive in the task queue. Also, the~number of rewards depends on the number of actions. Thereby, the~agent transitions cannot be stored in a sequential order, $( s_1, a_1, r_1, s_2, ..., s_T )$. This~violates the standard MDP assumption. The~bottom figure truncates the reward sequence in between the scheduling time step so that the agent receives the reward based on the onward task duration. In~this case, the~computing reward approximates the true reward value, but~the agent time step and environment time step become consistent. The~right figure shows a standard steady-state, which is when all jobs are stacked to the job queue and a pseudo-steady-state, which approximates the steady-state. In~a pseudo-steady-state, all jobs are stacked to the job queue without capturing previous decisions. This~disregards the past decisions but having a non-empty job~queue.}
    \label{fig:sched-timeline}
\end{figure}

In our environment, the~agent can observe the next state while executing scheduled tasks because any task having no predecessors can arrive at the task queue. As~illustrated in left diagram in Figure~\ref{fig:sched-timeline}, suppose a scheduled task at $t$ is completed at $\hat{t} \in [t+1, t+2]$. The~reward is received after task completion at $\hat{t}$, but~the next state can be received at $t+1$ due to the task dependency graph. Therefore, an~agent and environment time steps do not match, and~MDP transitions are not sequentially situated. More specifically, the~time step of an agent receiving a state and performing an action is different from the time step of an environment providing a state and a reward. In~particular, for~running with low scale values, the~injecting jobs are easily overlapped that giving additional complexity factors to the current~state.

To alleviate the problem, we~construct a reward function for the onward job duration, as~described in Equation~\eqref{eq:mean}. Since the reward function is computed based on the jobs that are currently executing, the~reward is continuously changing even when the previously scheduled task is not completed. We~truncate the reward sequence in between the agent scheduling time step so that the environment and agent become consistent with time step as shown the bottom of Figure~\ref{fig:sched-timeline}. The~reward refers to the ongoing jobs' duration, and~its sequence can be varied and prolonged depending on the previous action duration, as~specified in Equation~(\ref{eq:mean}). To~approximate the prolonged reward sequence, we~truncated the reward sequence as $\Tilde{r}_t = R_{t'}$, where $t'= \min(t, \hat{t})$. In~RL formulation, the~reward is a random variable induced by the selection of action. Hence, the~agent computes the return with the expectation of the cumulative rewards, and~the same return values can be used in the delayed reward case~\cite{schuitema2010control,katsikopoulos2003markov}. Moreover, we~add an extra ``no-op'' action when the ready task is not replenished to the ready queue. At~this time step, the~environment recalculates a reward and updates it to the rollout storage. This~produces an updated reward with delayed~action. 
  
Additionally, to~efficiently train the agent, we~present a `pseudo-steady-state' approximating operational conditions and train the agent using curriculum learning. Before~evaluating the scheduler performance, the~system starts from an empty job queue and injects jobs into the queue until it is filled compactly. As~illustrated in Figure~\ref{fig:jobgen-timeline}, we~empirically set a warm-up period, which is the time for the simulation to reach a steady state. For~training DeepSoCS, it is very time-consuming to wait for filling the job queue. Hence, before~running the environment, all jobs are injected into the job queue. We~refer to this state as `pseudo-steady-state', which approximates the~steady-state. 

\subsection{Joint~Action}
\label{sec:proposed-method:joint-action}
In multi-agent reinforcement learning, a~group of agents performs individual actions given a common state. One of the possible objectives is to receive a single high reward for joint action. In~our DeepSoCS architecture, as~we execute a task at the time of a given state, in~addition to delayed rewards, we~have an asynchronous reward for each task that is executed at a different time and computes its reward (based on its execution duration on a processing element) when the task finishes at a different time step. This~means we have multiple asynchronous task-based actions (of a single job-based action) that operate on a single, same state. In~other words, the~next state is computed by a stochastic combination of multiple, asynchronous task-based actions that approximate a single job-based action. The~rewards returned by the environment for the executions of task-based actions trigger stochastic gradient descents through the neural networks. The~joint action is approximated by multiple, asynchronous task-based actions based on the current state. The~result of the stochastic application of multiple asynchronous actions on the environment approximates the next state of the joint action. As~the tasks together form a job DAG, the~stochastic effects of task-based action is bounded by the fact that they are constrained by the underlying, constraining job, which is to say the state representation of a job inherently has a number of ready tasks. Specifically, as~task scheduling does not typically belong to an adversarial environment, which is the case of our DeepSoCS running in DS3 emulator, we~merely need to have monotonicity between greedy individual policies (of~associated individual, task-based actions) and greedy centralized or joint policy based on the optimal joint action-value function. Each action can execute in a decentralized manner entirely by its policy, choosing the greedy action to its Q-value. A~global argmax computation conducted on joint Q-value give the same expected result as a set of individual argmax computations carried out on each action's Q-value. DeepSoCS policies satisfy this monotonicity criterion as it chooses the smallest expected task execution latency for both individual actions and joint action. Formally, monotonicity is defined as a constraint on the relationship between each Q-value of individual action and the Q-value of the joint action, as~follows:
\begin{equation}
\label{eq:joint-action}
    \frac{\partial Q_{joint-action}}{\partial Q_{each-action}} >=  0  
\end{equation}

%%%%%%%%%%%%%%%%%%%%%%%%%%%%%%%%%%%%%%
%\input{experiments}
\section{Experiments}
\label{sec:experiments}
\textls[-15]{DS3 simulation continuously injects jobs throughout the simulation length. The~job is injected at every $t$ time step, where $t \sim \text{Exp}(\frac{1}{\text{scale}})$. The~lower the scale value, the~faster job injects to the job queue. We~empirically found that the injection speed exponentially increases between 100 scale and 50 scale. At~a 50 scale value, for~instance, 20 jobs are injected at every time step. Throughout the experiments, the~DS3 simulation allows stacking up to 12 jobs to the job queue. As~described in Section~\ref{sec:proposed-method:delayed-outcome}, the~warm-up period leads to steady-state condition. DeepSoCS uses pseudo-steady-state in the training phase. Table~\ref{table:exp-cond} provides the rest of the experiment settings. PSS refers to pseudo-steady-state, and~SS refers to~steady-state.}

\begin{table}[H] 
  \caption{Experiment~condition.}
  \label{table:exp-cond}
  \centering
  \begin{tabular}{lcccc}
    \toprule
    \textbf{Figure} & \textbf{Simulation Length} & \textbf{Warm-Up Period} & \textbf{Scale} & \textbf{PSS/SS} \\
    \midrule %MDPI: We change the format and add the link, please check.
    Figure~\ref{fig:top-wifi-res} (HEFT)     & 100,000 & 20,000 & -  & SS \\
    Figure~\ref{fig:top-wifi-res} (DeepSoCS) & 100,000 & 20,000 & -  & SS \\
    Figure~\ref{fig:gantt-single}            & 100,000 & 20,000 & 50 & SS \\
    Figure~\ref{fig:top-n-wifi-latency} (HEFT)     & 30,000  & 20,000 & 50 & SS \\
    Figure~\ref{fig:top-n-wifi-latency} (DeepSoCS) & 10,000  & 0      & 50 & PSS \\
    \bottomrule
  \end{tabular}
\end{table}

\begin{figure}[H]
  \centering
    \includegraphics[height=0.66\linewidth]{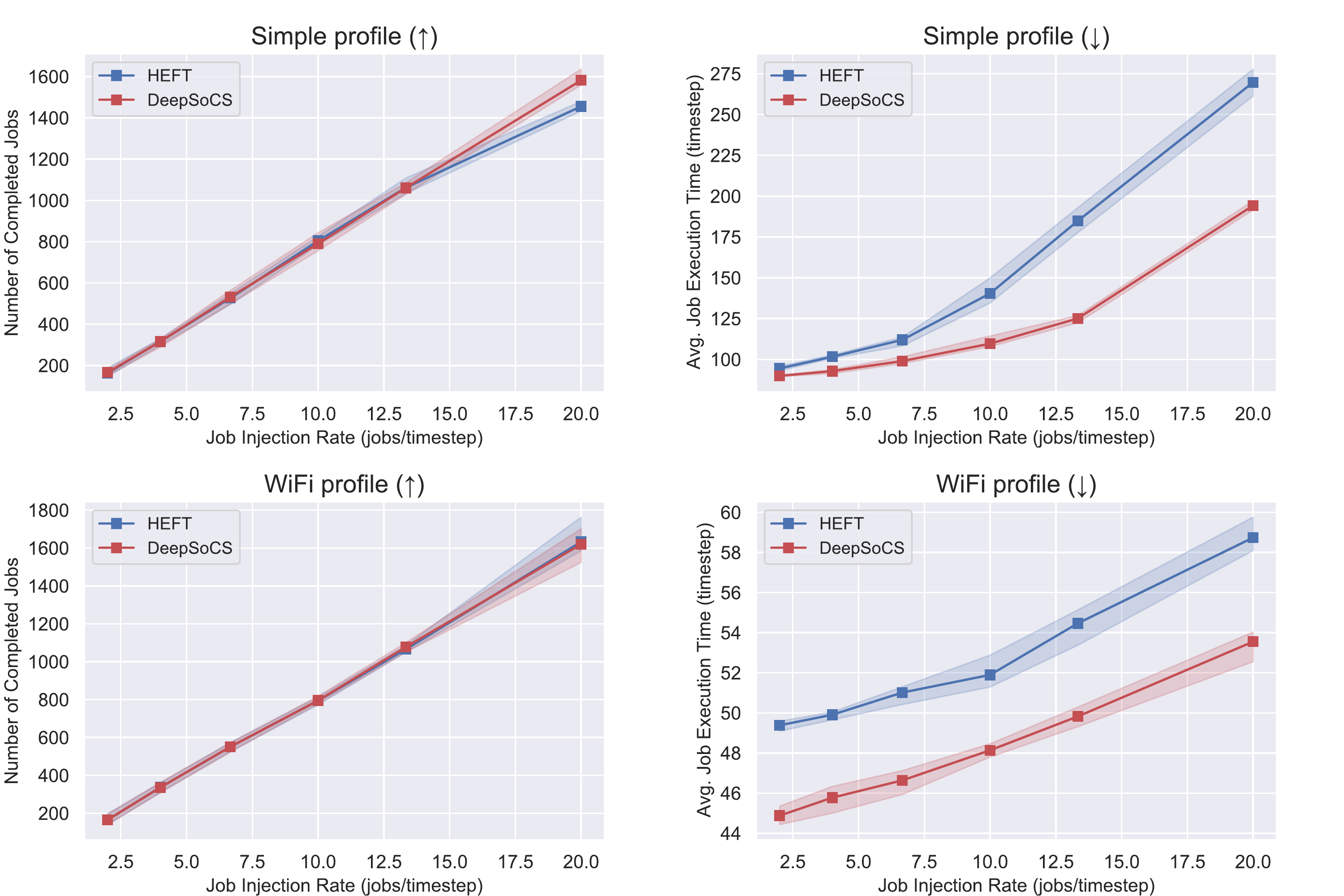}
    \caption{Figures show the average latency for DeepSoCSs and HEFTs with adding standard deviations. Left shows simple profile, and~right shows WiFi profile. Note that the HEFT shows poor performances after adding variations. All tested with scale of~50.}
    \label{fig:top-wifi-res}
\end{figure}

Figure~\ref{fig:top-wifi-res} shows performance evaluation with a canonical job profile~\cite{topcuoglu2002performance} and more complex file, a~WiFi profile which is described in Appendix~\ref{sec:appendix:wifi}. Each algorithm was tested on different scale values. We~ran 5 trials using different random seeds. The~x-axis represents the job injection rate. The~faster job injects as it goes to the right. Since the simulation allows stacking 12 jobs to the job queue at most, the~minimum scale of 50 is sufficient to validate rigorous test conditions. The~y-axis represents the number of completed jobs for the left plot, and~average latency for the right plot. For~the left plots, DeepSoCS and HEFT complete similar number of jobs in both simple and WiFi profiles. On~the other hand, DeepSoCS has smaller latency than HEFT. On~average, DeepSoCS peforms 7--9\% better than~HEFT.

% \begin{figure}[H]
  %\centering
   % \includegraphics[height=0.66\linewidth]{results/simple_wifi_perf.pdf}
    %\caption{The graph compares the results of DeepSoCs and HEFT on scheduling a canonical job profile (\textbf{top}) and a WiFi profile (\textbf{bottom}). The~arrow direction from the plot title shows better performance. The~description of WiFi profile is listed in Appendix~\ref{sec:appendix:wifi}.}
   % \label{fig:top-wifi-res}
%\end{figure}

\begin{figure}[H]
  \centering
    \includegraphics[height=0.3\linewidth]{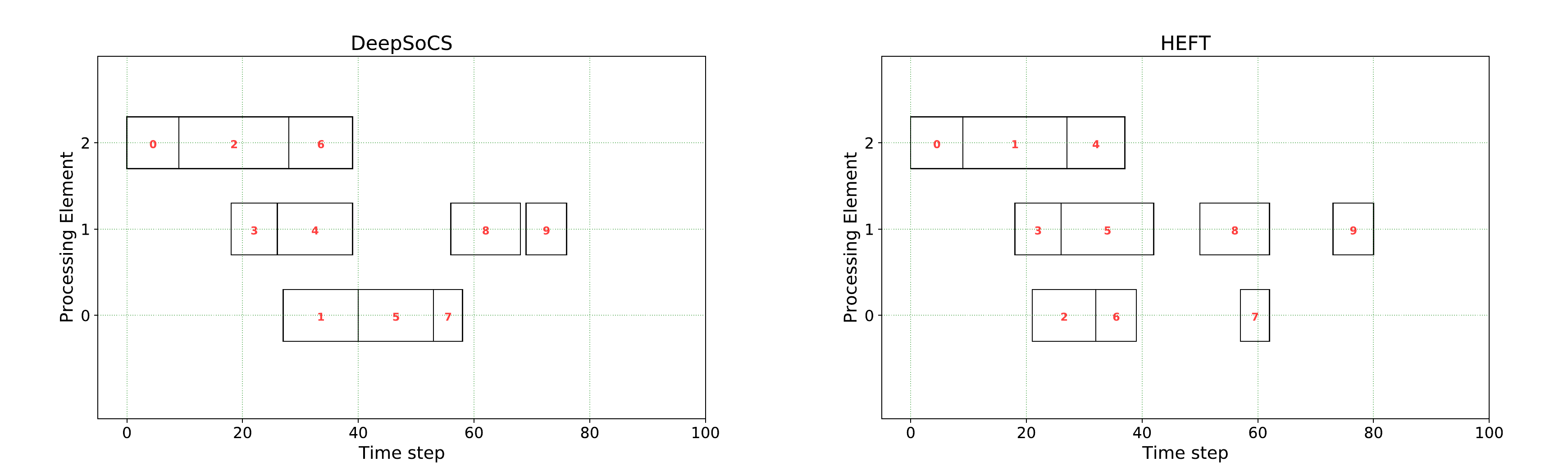}
    \caption{The Gantt charts for performing single job using DeepSoCS and HEFT algorithms are shown. A~simple profile has been~used.}
    \label{fig:gantt-single}
\end{figure}

To validate the outperformance, we~plotted the Gantt chart for DeepSoCS and HEFT in a simple profile. Figure~\ref{fig:gantt-single} shows a single input job injected with a scale of 50. Remark that both HEFT and DeepSoCS select PE using the same heuristic algorithm, and~the main difference is task prioritization. We~believe the reason behind this performance difference is that since HEFT greedily prioritize input tasks and map to designated tasks to PEs, the~algorithm potentially seeks myopic goals while, in~contrast, DeepSoCS trains via trial-and-error and its objective is to maximize the expected sum of rewards; therefore DeepSoCS has a more compact allocation in~total. 

\begin{figure}[H]
  \centering
    \includegraphics[height=0.33\linewidth]{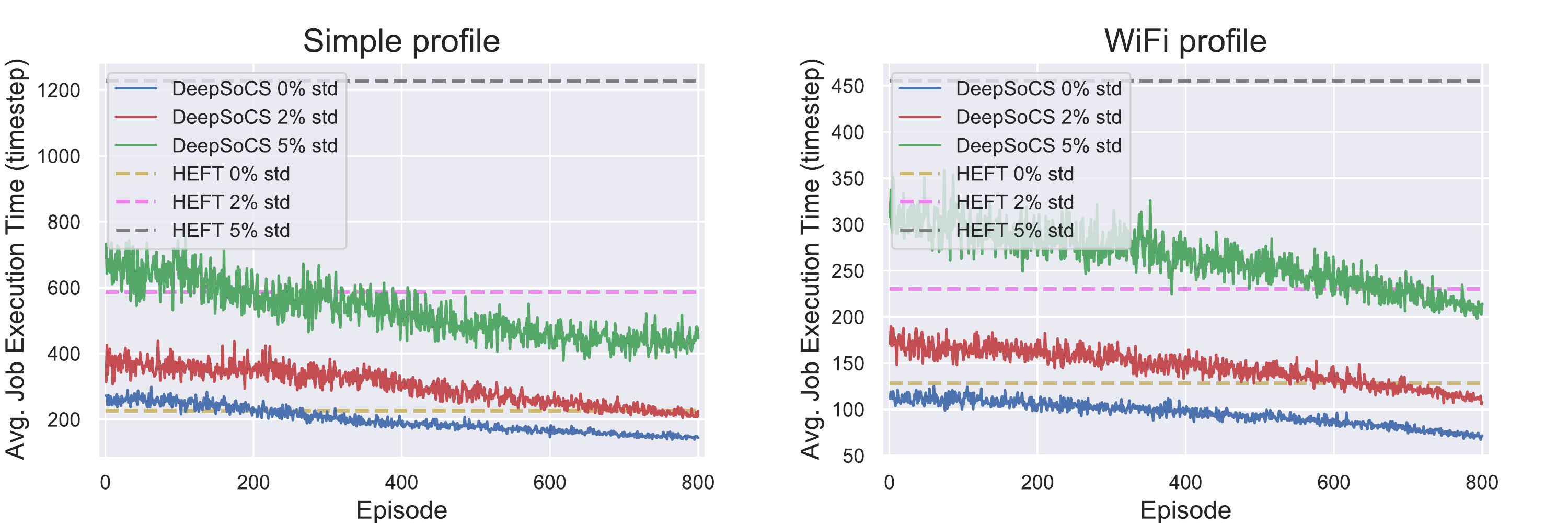}
    \caption{Figures show the average latency for DeepSoCSs and HEFTs with adding standard deviations. Left shows simple profile, and~right shows WiFi profile. Note that the HEFT shows poor performances after adding variations. All tested with scale of~50.}
    \label{fig:top-n-wifi-latency}
\end{figure}

\textls[-15]{In further experiments, we~consider uncertainty involved in the simulation. In~real-world application, PE performance can be perturbed by the thermal, physical malfunction, or~other environmental noises. Thus, we~add Gaussian noises to the supported functionalities in PEs and tested experiments as shown in Figure~\ref{fig:top-n-wifi-latency}.}

\textls[-15]{As described in Section~\ref{sec:introduction}, HEFT cannot capture stochastic PE performances and has no generalization} because the algorithm makes task orders based on the $rank_u$ values computed with a static resource profile. In~contrast, DeepSoCS shows stable performance even in noise added stochastic environments, and~performs with significantly lower latency compared to that of~HEFT.

\begin{figure}[H]
  \centering
    \includegraphics[height=0.4\linewidth]{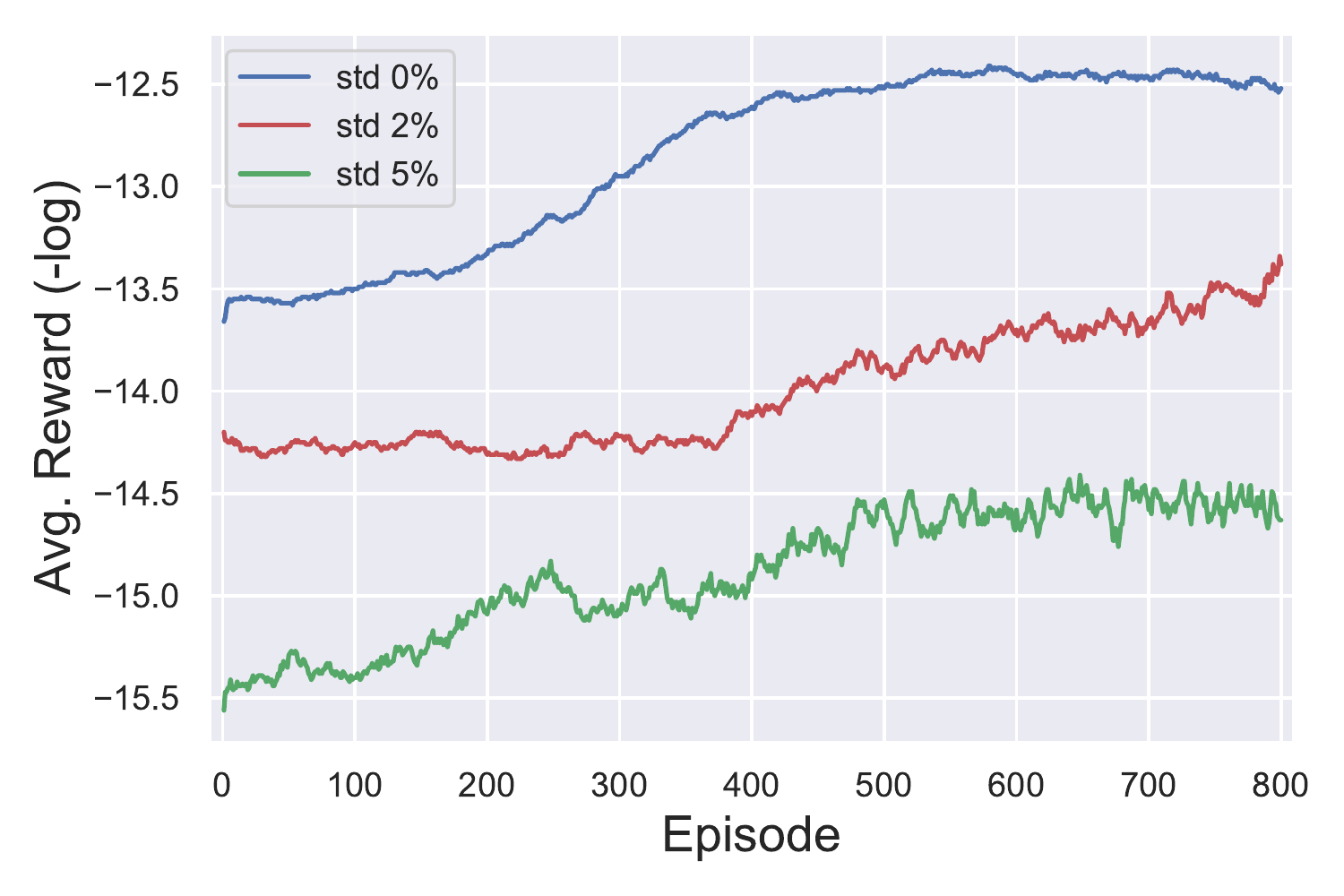}
    \caption{\textls[-22]{A figure shows cumulative rewards over training episodes in different PE variations. The~reward} scale is represented in negative log scale. The~agent starts training from pseudo-steady-state, and~uses simple input profile and 50 scale~setting.}
    \label{fig:reward-curve}
\end{figure}

In addition, Figure~\ref{fig:reward-curve} shows the cumulative reward curves for DeepSoCS with different variations to PE performances. In~this training phase, we~use scale of 50 for the most difficult problem setting, and~pseudo-steady-state to faster~training.

\unskip

%%%%%%%%%%%%%%%%%%%%%%%%%%%%%%%%%%%%%%
%\input{related}
\section{Related~work}
\label{sec:related-work}
\textls[-15]{There is a large body of work in reinforcement learning on scheduling or resource allocation problems. DRM first employs deep reinforcement learning to schedule a simple job resource allocation that does not have job hierarchy and homogeneous setting~\cite{mao2016resource}. Distributed Q-Learning has been used to schedule tasks to PEs in run-time~\cite{xiao2019self} with good results but only after preprocessing steps of compiling an application code into Instruction Dependency Graph and forming task pools via compile-time resource allocation via neural network classifiers and community detection. QL-HEFT combines Q-learning and HEFT to show better performance when increasing the number of tasks~\cite{tong2019ql}. However, it used tabular Q-learning and did not consider joint action and overlapping jobs. In~general, HEFT-based methods are capable of finding approximate solutions for NP-hard scheduling problems but are restricted by expert's static global point of view and domain knowledge of task scheduling vis-a-vis dynamic, fine-grained realities of task scheduling where jobs own many tasks, and~they can overlap with one another. It does not consider overlapping and continuously injecting jobs, which is not an ideal problem setting in DS3. Also, QL-HEFT uses HEFT's $rank_u$ value to the positive reward function, which is not appropriate for scheduling applications where reducing the amount of execution time is a critical metric. ADTS presents Monte-Carlo Tree Search (MCTS) with a policy gradient-based REINFORCE agent for static DAG tasks scheduling but not for dynamic DAG nor overlapping jobs~\cite{cheng2019smart}. SCARL architecture employs attentive embedding~\cite{vaswani2017attention} to schedule jobs to heterogeneous multi-resource cluster~\cite{cheong2019scarl}. In~its work, the~input data type is relatively simple, which has one-level and a static~structure. }

The task and PE association is closely related to the combinatorial problem. As~an example, the~device placement selects hardware modules to individual layers from large neural networks. RL-based placement incorporates the sequence-to-sequence model and REINFORCE algorithm to address device optimization~\cite{mirhoseini2017device,mirhoseini2018hierarchical}. Placeto generalizes device placement in any computation graph leveraging graph embeddings~\cite{addanki2019placeto}. Deep Reinforcement Relevance Network addresses combinatorial action spaces in natural language processing applications by forwarding both state and action embeddings to the networks~\cite{he2016deep}. Branching Dueling Q-Network was developed with action branching architecture to handle discrete joint-action and experimented with physical simulator~\cite{tavakoli2018action}. S2V-DQN uses Structure2Vec and Q-learning to address various combinatorial problems~\cite{khalil2017learning}. Subsequently, the attention model with REINFORCE algorithm addresses routing optimization problems~\cite{kool2018attention}. From~the perspective of all possible combinations of joint actions, Wolpertinger Architecture uses Wolpertinger Policy leveraging k-nearest neighbors and proto-action value function to address large action spaces~\cite{dulac2015deep}. Multi-agent reinforcement learning based on DQN finds correlated equilibrium between makespan and cost for workflow scheduling in a Markov game setting with joint action and joint state.~\cite{Wang2019multi}.

%%%%%%%%%%%%%%%%%%%%%%%%%%%%%%%%%%%%%%
%\input{conclusion}
\section{Conclusions}
\label{sec:conclusion}
In this paper, we~present a novel neural network algorithm DeepSoCS that learns to make the extremely resource-efficient task ordering actions in the high-fidelity environment. With~two novel neural network designs, hierarchical job- and task-graph embeddings, and~efficient use of real-time task information in the state space, DeepSoCS is capable of learning hierarchical job scheduling to heterogeneous resources. Also, DeepSoCS solves delayed consequences and joint-action that arise from applying DRL to the highly realistic environment by using reward shaping and new joint-action formalization. We~empirically show that DeepSoCS demonstrates the robustness and system-wide performance gains in job execution time under realistic noise conditions over~HEFT.

As mentioned in Sections~\ref{sec:introduction} and~\ref{sec:prob-def:challenges}, the~observation state does not fully represent the overlapping jobs with a continually changing environment. We~consider the problem as a partially observable Markov decision process. To~resolve uncertainty in the states, we~plan to add temporal information such as Long Short-Term Memory (LSTM)~\cite{hochreiter1997long} to the model. Alternatively, leveraging HEFT experience to train neural networks~\cite{hester2018deep} may speed up the training time and further improve its performance. To~process task and PE selections end-to-end, we~need to resolve combinatorial complexity in the PE manager. Thereby, we~expect to apply an attention-based model, which has permutation-invariance property, to~the PE manager~\cite{kool2018attention}. To~improve DeepSoCS into a more practical algorithm, scheduling heterogeneous profiled jobs and reducing both execution time and power consumption is a very promising algorithm for scheduling~applications.

%%%%%%%%%%%%%%%%%%%%%%%%%%%%%%%%%%%%%%
\vspace{6pt} 

%%%%%%%%%%%%%%%%%%%%%%%%%%%%%%%%%%%%%%%%%%
%% optional
%\supplementary{The following are available online at \linksupplementary{s1}, Figure S1: title, Table S1: title, Video S1: title.}


%%%%%%%%%%%%%%%%%%%%%%%%%%%%%%%%%%%%%%%%%%
\authorcontributions{Conceptualization, T.T.S., A.Y. and B.R.; Funding acquisition, C.-B.S.; Investigation, T.T.S. and A.Y.; Methodology, T.T.S., J.H., J.K. and A.Y.; Project administration, B.R.; Software, T.T.S. and J.H.; Supervision, B.R.; Writing---original draft, T.T.S. and J.K.; Writing---review \& editing, J.K., A.Y. and B.R. All authors have read and agreed to the published version of the manuscript.}

%%%%%%%%%%%%%%%%%%%%%%%%%%%%%%%%%%%%%%%%%%
\funding{APC was funded by the MSIT (Ministry of Science and ICT), Korea, under the ITRC (Information Technology Research Center) support program (IITP-2020-2016-0-00288) supervised by the IITP (Institute for Information \& communications Technology Planning \& Evaluation)}

%%%%%%%%%%%%%%%%%%%%%%%%%%%%%%%%%%%%%%%%%%
% \acknowledgments{}

% %%%%%%%%%%%%%%%%%%%%%%%%%%%%%%%%%%%%%%%%%%
\conflictsofinterest{The authors declare no conflicts of interest.}

%%%%%%%%%%%%%%%%%%%%%%%%%%%%%%%%%%%%%%%%%%
%% optional
% \abbreviations{The following abbreviations are used in this manuscript:\\

% \noindent 
% \begin{tabular}{@{}ll}
% MDPI & Multidisciplinary Digital Publishing Institute\\
% DOAJ & Directory of open access journals\\
% TLA & Three letter acronym\\
% LD & linear dichroism
% \end{tabular}}

%%%%%%%%%%%%%%%%%%%%%%%%%%%%%%%%%%%%%%%%%%
%% optional
\appendixtitles{yes} %Leave argument "no" if all appendix headings stay EMPTY (then no dot is printed after "Appendix A"). If~the appendix sections contain a heading then change the argument to "yes".
\appendix
\section{WiFi Job~Profile}\label{sec:appendix:wifi}

\begin{figure}[H]
  \centering
    \includegraphics[height=0.4\linewidth]{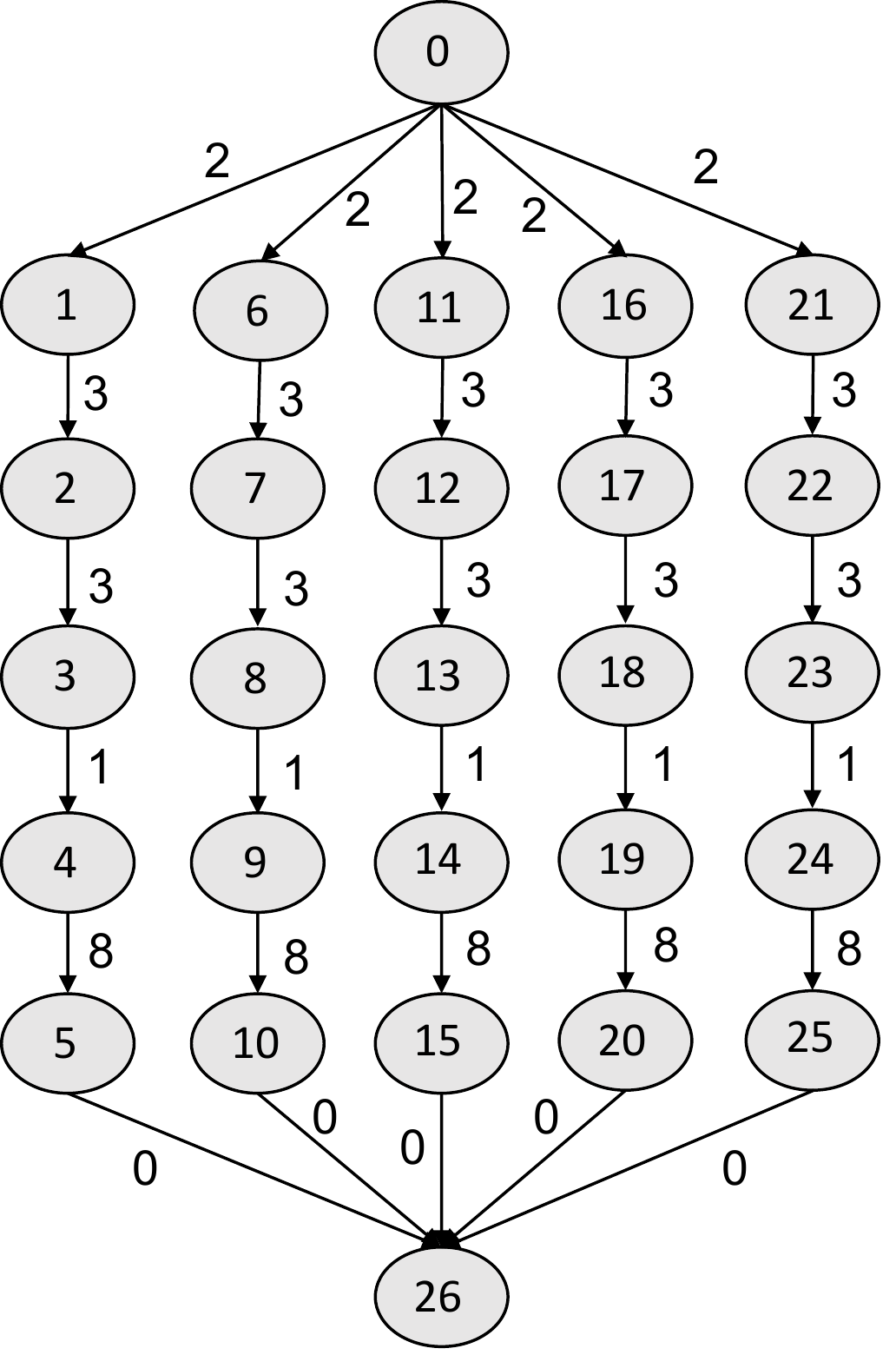}
    \caption{A more complicated WiFi job profile. The~job has 25~tasks.}
    \label{fig:wifi-profile}
\end{figure}

\textls[-15]{The more complicated WiFi profile is shown in Figure~\ref{fig:wifi-profile} and Table~\ref{table:wifi-comm}. Comparing to the canonical profile, WiFi has 7 different types of resources. In~a resource file, 4 type 1s, 4 type 2s, 1 type 3, 1 type 4, 2~type 4s, 2 type 5s, and~3 type 5s. Note that the task 4, 9, 14, 19, and~24 has very high communication cost, but~if the scheduler takes type 5 or 6, the~execution time would be significantly reduced. The~resource type 3 to 7 have unsupported functionalities and need to be extra careful for selecting~resources.}

\vspace{-6pt}

\begin{table}[H]
  \caption{A more complicated WiFi resource profile. There are 7 different types of~resources.}
  \label{table:wifi-comm}
  \centering
\scalebox{0.9}[0.9]{
  \begin{tabular}{cccccccc}
    \toprule
    \textbf{Task} & \textbf{Type 1} & \textbf{Type 2} & \textbf{Type 3} & \textbf{Type 4} & \textbf{Type 5} & \textbf{Type 6} & \textbf{Type 7} \\
    \midrule
    0    &   10   &   22   &   2   &   1   &   -   &   -   &   -   \\
    1    &   4    &   22   &   -   &   -   &   -   &   -   &   -   \\
    2    &   8    &   22   &   -   &   -   &   -   &   -   &   -   \\
    3    &   3    &   22   &   -   &   -   &   -   &   -   &   -   \\
    4    &   118  &   296  &   -   &   -   &   3   &   2   &   -   \\
    5    &   3    &   5    &   2   &   1   &   -   &   -   &   -   \\
    6    &   4    &   10   &   2   &   1   &   -   &   -   &   -   \\
    7    &   8    &   15   &   2   &   1   &   -   &   -   &   -   \\
    8    &   3    &   5    &   2   &   1   &   -   &   -   &   -   \\
    9    &   118  &   296  &   -   &   -   &   3   &   2   &   -   \\
    10   &   3    &   5    &   2   &   1   &   -   &   -   &   -   \\
    11   &   4    &   10   &   2   &   1   &   -   &   -   &   -   \\
    12   &   8    &   15   &   2   &   1   &   -   &   -   &   -   \\
    13   &   3    &    5   &   2   &   1   &   -   &   -   &   -   \\
    14   &   118  &   296  &   -   &   -   &   3   &   2   &   -   \\
    15   &   3    &   5    &   2   &   1   &   -   &   -   &   -   \\
    16   &   4    &   10   &   2   &   1   &   -   &   -   &   -   \\
    17   &   8    &   15   &   2   &   1   &   -   &   -   &   -   \\
    18   &   3    &   5    &   2   &   1   &   -   &   -   &   -   \\
    19   &   118  &   296  &   -   &   -   &   3   &   2   &   -   \\
    20   &   3    &   5    &   2   &   1   &   -   &   -   &   -   \\
    21   &   4    &   10   &   2   &   1   &   -   &   -   &   -   \\
    22   &   8    &   15   &   2   &   1   &   -   &   -   &   -   \\
    23   &   3    &   5    &   2   &   1   &   -   &   -   &   -   \\
    24   &   118  &   296  &   -   &   -   &   3   &   2   &   -   \\
    25   &   3    &   5    &   2   &   1   &   -   &   -   &   -   \\
    \bottomrule
  \end{tabular}}
\end{table}

% \unskip
% \subsection{}
% The appendix is an optional section that can contain details and data supplemental to the main text. For example, explanations of experimental details that would disrupt the flow of the main text, but nonetheless remain crucial to understanding and reproducing the research shown; figures of replicates for experiments of which representative data is shown in the main text can be added here if brief, or as Supplementary data. Mathematical proofs of results not central to the paper can be added as an appendix.

%%%%%%%%%%%%%%%%%%%%%%%%%%%%%%%%%%%%%%%%%%

%\externalbibliography{yes}
%\bibliography{ref}

\reftitle{References}

%%%%%%%%%%%%%%%%%%%%%%%%%%%%%%%%%%%%%%%%%%
\end{document}